\documentclass[journal,twoside,web]{ieeecolor}
\usepackage{tmi}
\usepackage{cite}
\usepackage{amsmath,amssymb,amsfonts}
\usepackage{algorithmic}
\usepackage{graphicx}
\usepackage{textcomp}
\usepackage{multirow}
\def\BibTeX{{\rm B\kern-.05em{\sc i\kern-.025em b}\kern-.08em
    T\kern-.1667em\lower.7ex\hbox{E}\kern-.125emX}}
\begin{document}
\title{Multi-Level Global Context Cross Consistency Model for Semi-Supervised Ultrasound Image Segmentation with Diffusion Model}
\author{Fenghe Tang, Jianrui Ding, Lingtao Wang, Min Xian, Chunping Ning. 
\thanks{This work is supported by Shandong Natural Science Foundation of China (ZR2020MH290) and by the Joint Funds of the National Natural Science Foundation of China (U22A2033).\textit{(Corresponding author: Jianrui Ding.)} }
\thanks{Fenghe Tang, Jianrui Ding and Lingtao Wang are with the School of Computer Science and Technology, Harbin Institute of Technology, Harbin , China 150001 (e-mail: 543759045@qq.com; jrding@hit.edu.cn; 453260638@qq.com). }
\thanks{Min Xian is with Department of Computer Science, University of Idaho, Idaho Falls, ID 83401, USA (e-mail: mxian@uidaho.edu).}
\thanks{Chunping Ning is with the Ultrasound Department, The Affiliated Hospital of Qingdao University, Qingdao, China. (e-mail: 152081340@qq.com).}}

\maketitle
\begin{abstract}
Medical image segmentation is a critical step in computer-aided diagnosis, and convolutional neural networks are popular segmentation networks nowadays. However, the inherent local operation characteristics make it difficult to focus on the global contextual information of lesions with different positions, shapes, and sizes. Semi-supervised learning can be used to learn from both labeled and unlabeled samples, alleviating the burden of manual labeling. However, obtaining a large number of unlabeled images in medical scenarios remains challenging. To address these issues, we propose a Multi-level Global Context Cross-consistency (MGCC) framework that uses images generated by a Latent Diffusion Model (LDM) as unlabeled images for semi-supervised learning. The framework involves of two stages. In the first stage, a LDM is used to generate synthetic medical images, which reduces the workload of data annotation and addresses privacy concerns associated with collecting medical data.  In the second stage, varying levels of global context noise perturbation are added to the input of the auxiliary decoder, and output consistency is maintained between decoders to improve the representation ability. Experiments conducted on open-source breast ultrasound and private thyroid ultrasound datasets demonstrate the effectiveness of our framework in bridging the probability distribution and the semantic representation of the medical image. Our approach enables the effective transfer of probability distribution knowledge to the segmentation network, resulting in improved segmentation accuracy. The code is available at https://github.com/FengheTan9/Multi-Level-Global-Context-Cross-Consistency.
\end{abstract}


\section{Introduction}
\label{sec:introduction}
\IEEEPARstart{U}{LTRASOUND} imaging is a non-invasive, radiation-free and cost-effective diagnostic modality that has been widely utilized in the detection of thyroid nodules, breast tumors, and gonadal tissues [1]. However, ultrasound images often suffer from noise, low contrast and resolution, which makes manual analysis a time-consuming and laborious process. Furthermore, subjective factors such as the experience and mental state of radiologists can lead to misdiagnosis. Automated medical image analysis can effectively overcome these limitations. Deep learning technology has made significant progress in various automated medical image analysis tasks, especially in medical image segmentation using convolutional neural networks (CNNs). However, training a robust model requires a large amount of labeled data, which is challenging in the case of medical images due to patient privacy protection and the need for domain experts for accurate annotation. In addition, ultrasound images can exhibit significant variability in appearance and shape due to differences in patient anatomy, imaging equipment, and imaging protocols. Therefore, performing medical image segmentation with a limited number of high-quality labeled ultrasound images is a challenging task.

Recently, there has been significant research focused on training robust models with limited annotated images, which can be broadly categorized into two approaches: weakly-supervised learning and semi-supervised learning (SSL). Weakly-supervised learning involves using annotated images that are easy to collect but imprecise for model training, whereas SSL leverages a limited set of labeled images along with a large number of unlabeled images to train a powerful network. However, in some scenarios, the variation in style and content of images acquired from different ultrasound devices can negatively impact the effectiveness of SSL. Furthermore, obtaining a substantial amount of unlabeled medical images from various sources can prove to be unfeasible due to privacy concerns.

Fortunately, generative models offer a practical solution to the challenges mentioned above by generating a significant number of synthetic images from existing ones [3]. This approach effectively addresses ethical concerns surrounding the acquisition of future images [4] while also improving the performance of target tasks [5], [6]. While generative adversarial models (GANs) [7] have been extensively used for image generation, they suffer from limited pattern coverage and cannot effectively capture the true diversity of features. Moreover, GANs are prone to training instability and mode collapse [8]. Currently, the Denoising Diffusion Probability Model (DDPM) [9] and Latent Diffusion Model [10] are the most advanced generative models for image generation tasks, owing to their exceptional pattern coverage and quality of the generated samples [11]. In this study, we focus on using a large number of synthetic samples as unlabeled data for semi-supervised learning. This approach not only significantly reduces the burden of data labeling, but also overcomes the obstacles posed by collecting a large amount of unlabeled private data. Moreover, we believe that synthetic samples generated by LDM possess target domain probability distribution knowledge that is absent in CNNs, and we establish a connection between them using semi-supervised learning.

To effectively leverage both labeled and unlabeled synthetic and real data, while also exploring the impact of the relationship between labeled and unlabeled data on semi-supervised models, we propose a multi-level global context cross-consistency framework (MGCC), as shown in \textcolor{cyan}{Figure 1}. In the first stage, we employ LDM to generate a substantial quantity of valuable synthetic samples from existing medical image data. In the second stage, inspired by ConvMixer [12] and cross-consistency training [13], we propose a fully convolutional semi-supervised medical image segmentation network with multi-level global context cross-consistency to effectively exploit the unlabeled synthetic samples.

To solve the limitation of ordinary convolution locality in fully convolutional networks, Trockman et al. proposed ConvMixer [12] which uses large convolutional kernels to mix remote spatial locations to obtain global context information.  Compared to the Transformer, ConvMixer, which employs convolutional inductive bias, is better suited for computer vision tasks and has a lower computational overhead than the self-attention mechanism.

To leverage unlabeled synthetic and real data and improve the model's robustness to various noise perturbations in global context information, we introduce different level global context noise perturbation to the decoders and maintain consistency among multiple decoding outputs. Specifically, we introduce different length of ConvMixer with perturbation in shared encoder to obtain different level global context noise perturbation. And we use them as inputs for multiple auxiliary decoders and maintain consistency between the main decoder and auxiliary decoders outputs.

Moreover, to suppress irrelevant features and enhance valuable encoder knowledge transfer, we propose a multi-scale attention gate in skip-connection stage to select significant features using different receptive fields. Our main contributions are as follows:

\begin{itemize}
\item
We propose a novel semi-supervised medical image segmentation framework that utilizes diffusion models to generate synthetic medical ultrasound images and employs these generated images as unlabeled data for semi-supervised learning. By utilizing synthetic samples generated by LDM, we address the challenges faced by collecting large amounts of unlabeled medical privacy samples. And by leveraging semi-supervised learning, we establish a connection between the diffusion probability distribution of the target domain and semantic representations, enabling effective transfer of diffusion probability distribution knowledge from the target domain to the segmentation network. To our knowledge, this work is the first to utilize data generated by diffusion models as unlabeled samples for semi-supervised learning.
\item
We propose a fully convolutional semi-supervised medical segmentation network with multi-level global context cross-consistency, where different levels of global context noise perturbation are introduced to the auxiliary decoder while maintaining consistency among decoder outputs. This approach improves the network’s representational ability for segmented objects with varying positions and significant morphological differences.
\item
Experiments on public medical datasets demonstrate the effectiveness of the proposed semi-supervised methods and confirm that our proposed method can maximize the ability of the segmentation model to learn diffusion probability knowledge.
\end{itemize}

This work is an extension of our previous work published on ISBI-2023 [14]. Unlike our previous work, we significantly expand our proposed network to enable semi-supervised learning and investigate the feasibility of utilizing synthetic samples from LDM as unlabeled data.

\section{RELATED WORK}
\subsection{Medical Image Generation}
Previous studies have utilized generative adversarial models (GANs) [15] to generate synthetic medical images [16]-[18]. However, due to their architecture, GANs have limitations in generating high-resolution medical images with diverse internal details. Recently, the Denoising Diffusion Probability Model (DDPM) [9], [10] has gained considerable attention for addressing various deep learning generation problems [19]. DDPM's flexible model architecture and ability to perform exact log-likelihood computations make it a promising alternative to GANs.

Despite the impressive achievements of DDPM in generating various medical images, such as MRI [20], ophthalmology, lung CT, histopathology [6], and genomic information images [21], there has been limited research on utilizing DDPM for ultrasound image generation. In this study, we utilize the Latent Diffusion Model (LDM) [10] to generate ultrasound synthesis images. Compared to traditional DDPM, LDM substantially reduces the computational and inference costs while generating high-resolution ultrasound images. 

\subsection{Medical Image Segmentation}
Convolutional neural networks (CNNs) have gained popularity in medical image segmentation due to their powerful deep learning capabilities, and among the various CNN variants, U-Net [22] stands out for its superior performance. U-Net is a pyramid-structured segmentation network based on an encoder-decoder architecture that transfers semantic information between the encoder and decoder through skip-connections. In recent years, several U-Net based medical segmentation networks have been proposed, such as U-Net++ [23], Attention U-Net [24], Unet3+ [25] and UNeXt [26].

The limitation of the convolution operation is its restricted ability in capturing global contextual information, which is essential for accurate medical image segmentation. Recently, several networks based on the Transformer model [27] have been applied to medical image segmentation [28]-[30] due to their ability to effectively extract global information from images. TransUnet [28] employs Vit [31] to obtain global context with CNN, but it requires massive medical images and computing overhead. Compared to the hybrid CNN and Transformer structure, we believe that a fully convolutional network can still perform medical ultrasound image segmentation efficiently and effectively. In our previous work, we proposed a fully convolutional medical image segmentation network, CMU-Net [14], which uses ConvMixer instead of Transformer to extract global context information.
\subsection{Semi-Supervised Medical Image Segmentation}
Manual pixel-wise labeling by medical professionals is a time-consuming task. SSL based medical segmentation methods [13], [32]-[38] utilize a large amount of unlabeled data and a small number of high-quality labeled data to alleviate the labeling burden on medical professionals. Previous work on semi-supervised segmentation can be categorized into two major approaches: pseudo-label-based iterative learning [32] and consistency-based joint training [33]. Tarvainen and Valpola [33] proposed the mean teacher (MT) model, which leverages supervised loss and consistency loss for labeled and unlabeled data, respectively. Additionally, Yu et al. [34] applied the uncertainty estimation of Monte Carlo Dropout to the MT model. Chen et al. [37] performed cross-pseudo supervision through two differently initialized networks of the same structure. Luo et al. [38] extended it to cross-teaching between CNN and Transformer. Furthermore, Luo et al. [38] built consistency by performing both segmentation and level set regression tasks. Ouali et al. [13] proposed a segmentation method based on cross-consistency training, which employs multiple perturbed encoder feature as auxiliary decoder inputs and ensures decoder output consistency to enhance the encoder's representations.

Most existing semi-supervised segmentation methods use U-Net as the backbone, but due to the locality limitation of ordinary convolution, these networks cannot effectively extract global context. Although Luo et al. [37] introduced Transformer into the framework for cross teaching, training a powerful self-attention mechanism network requires a significant amount of data and computational resources. In addition, even though SSL reduces the time and labor involved in labeling, it still faces challenges in obtaining a large number of unlabeled images in medical scenarios. In contrastto traditional SSL, we use generated images as unlabeled samples and introduce ConvMixer in SSL to achieve multi-level global context consistency. We also establish a link between representation learning and generative learning and leverage SSL to obtain the diffusion probability distribution knowledge.

\section{METHODOLOGY}
\subsection{Overview}
Our proposed multi-level global context cross-consistency framework is illustrated in \textcolor{cyan}{Figure 1} and consists of two steps: (1) Medical image synthesis: Leverage the LDM to generate a large number of valuable synthetic samples. (2) Multi level global context cross-consistency network: To utilize synthesized unlabeled images, we input different levels of global context noise perturbation to different auxiliary decoders and maintain consistency between the output of the main and auxiliary decoders. We aim to establish the network's ability to produce consistent segmentation results on varying levels of global contextual noise perturbations.

In this work, the training set includes three subsets: labeled dataset $D_N^l$ with N annotated samples, unlabeled dataset $D_M^u$ with M unannotated samples and unlabeled synthetic dataset $D_A^u$ with A unannotated images which generated by LDM from $D_N^l$ and $D_M^u$. Thus, the entire unlabeled training set can be denoted as $D_{M+A}^u=D_M^u\cup D_A^u$ and the entire training set is $D_{N+M+A}=D_N^l\cup D_M^u\cup D_A^u$. We presuming that an image is $x_i$, $x_i$ has ground truth $y_i$ if $x_i\in D_N^l$. On the contrary, if $x_i\in D_{M+A}^u$, its label does not exist.

\subsection{Medical Ultrasound Image Generation Based on Latent Diffusion Model}
The traditional diffusion model operates in pixel space and requires significant computational resources and time for training. The LDM [10] is utilized to generate latent space codes, which are then decoded into pixel space, significantly reducing computational and inference costs while producing high-resolution images. The implementation of the LDM is divided into two parts: Pixel-level Image Compression and Latent Denoising Diffusion.
\subsubsection{Pixel-level Image Compression}
Pixel-level image compression involves training an encoder $\mathcal{E}$ to generate latent codes that correspond to the images. Additionally, the decoder $\mathcal{D}$ restores the latent codes to a high-resolution image. As shown in formula (1):
\begin{equation}z=\mathcal{E}(x) \   \hat{x}=\mathcal{D}(z)\label{eq}\end{equation}
where $x\in D_{N+M+A}$ represents the input image, $z$ represents the latent space code corresponding to $x$, and $\hat{x}\in D_A^u$ represents the synthesized high-resolution image obtained by the decoder.
\begin{figure}[b]
\centerline{\includegraphics[width=\columnwidth]{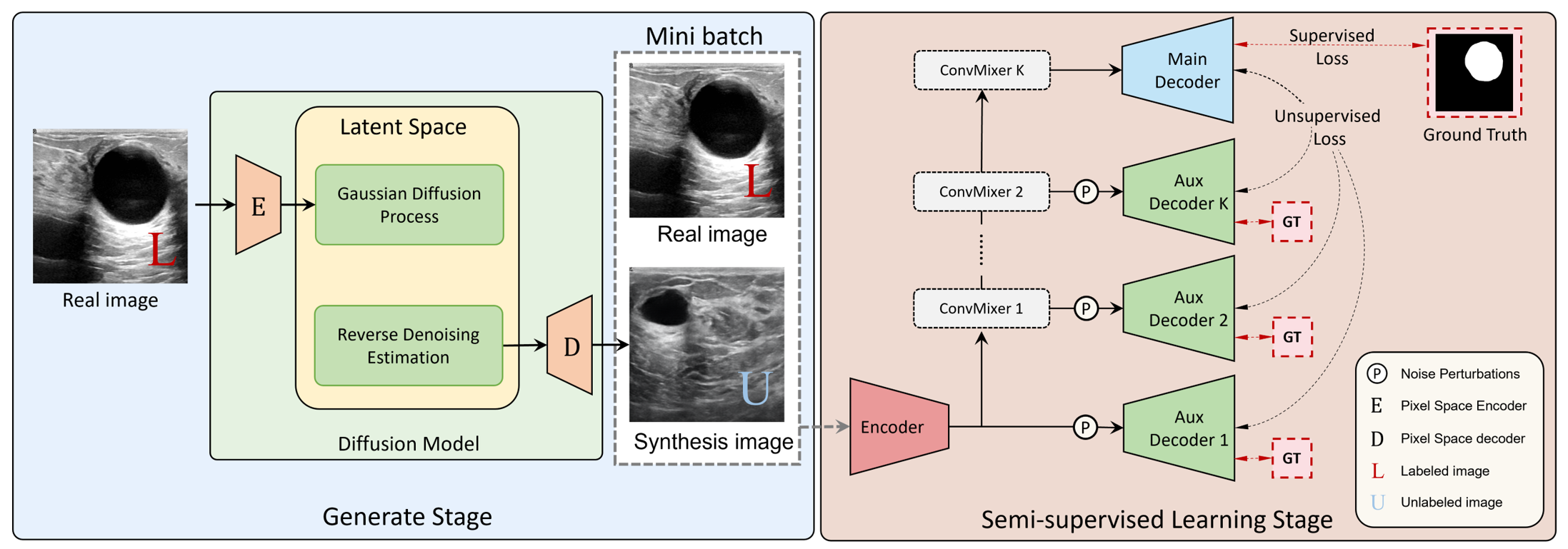}}
\caption{Overview of our proposed MGCC framework.}
\label{fig1}
\end{figure}

\begin{figure*}[t]
\centerline{\includegraphics[width=15cm]{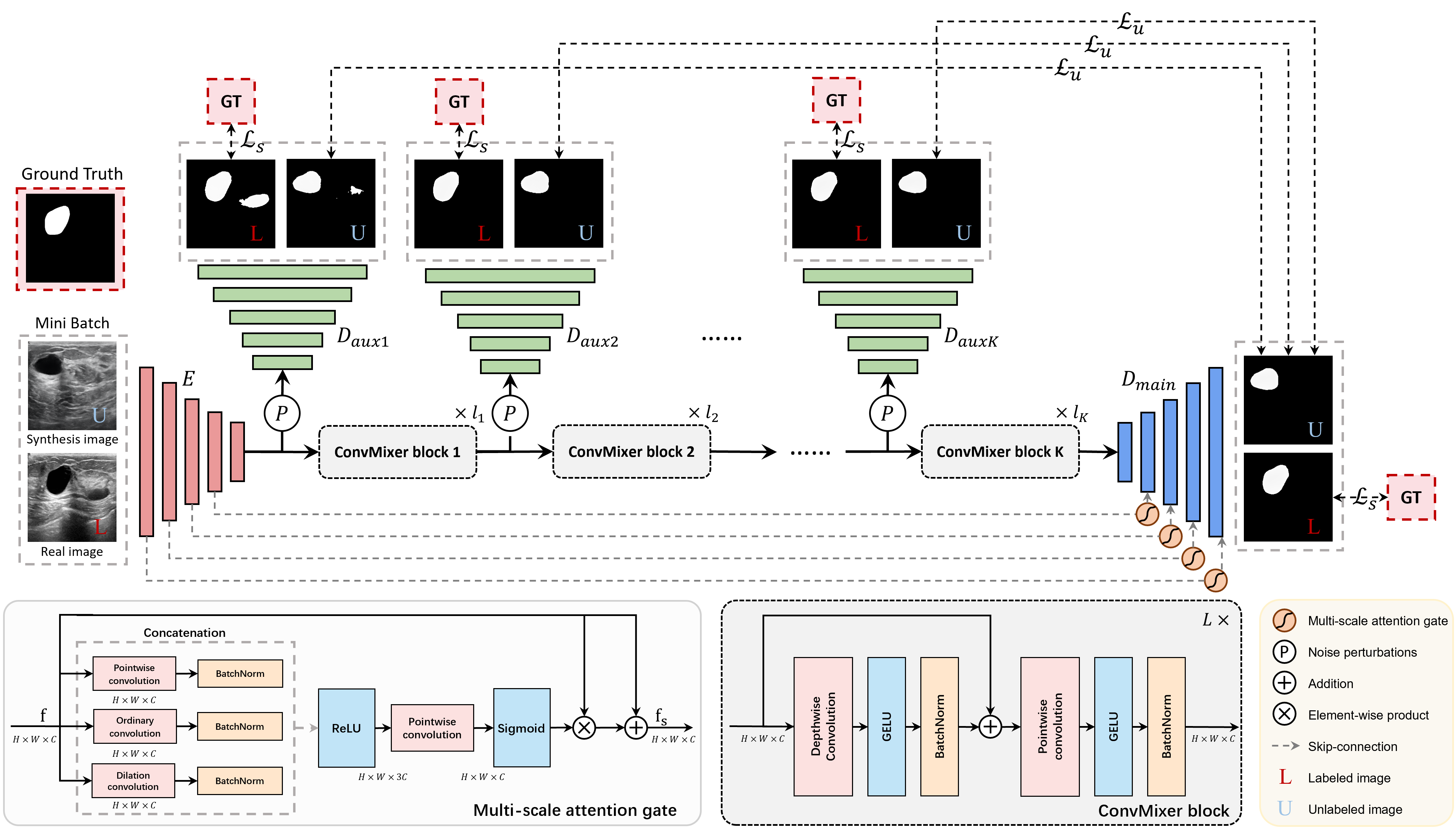}}
\caption{Overview of our proposed MGCC network.}
\label{fig2}
\vspace{-0.5cm}
\end{figure*}
\subsubsection{Latent Denoising Diffusion}
The latent diffusion model diffuses the latent space code by adding Gaussian noise while estimating the data distribution $p(x)$ by gradually denoising the Gaussian noise. The entire denoising learning process corresponds to the inverse process of a fixed Markov chain of length T. The loss function of LDM can be formulated as formula (2).

\begin{equation}\mathcal{L}_{LDM}=\mathbb{E}_{z,\epsilon\sim\mathcal{N}\left(0,1\right),t}[\vert\vert \epsilon-\epsilon_{\theta}(z_{t},t)\vert\vert_{2}^{2}]\label{eq}\end{equation}
where $\epsilon_{\theta}(z_{t},t), t= 1...T$ represents the denoising autoencoder which is used to predict the denoising distribution variable of $z_{t}$, and $z_{t}$ is the result of diffusing latent space code $z_{0}$ with $t$ times.

\subsection{Multi-Level Global Context Cross Consistency}
To implement multi-level global context cross-consistency learning for ultrasound image segmentation, we integrate multi-scale attention gates and a ConvMixer module into our proposed framework. The architecture is illustrated in \textcolor{cyan}{Figure 2}, which consists of a shared encoder E, a main decoder $D_{main}$, and K auxiliary decoders $\left\{D_{aux1},\ {\ D}_{aux2},\ldots,D_{auxK}\right\}$. We adopt the encoder-decoder structure of U-Net [22] in our work. In addition, we embed ConvMixer block of varying lengths between a shared encoder and multiple decoders to extract different levels of global context information. Furthermore, we integrate multi-scale attention gates with the skip-connection to enhance the efficient transfer of valuable encoder features. 

\subsubsection{ConvMixer Module and Multi-scale Attention Gates}
In this section, we present the implementation details of the ConvMixer block and multi-scale attention gates employed in our method. 

{\bf ConvMixer Module:}  The ConvMixer module [12] utilizes large convolution kernels to mix remote spatial locations to obtain global context information. As shown in \textcolor{cyan}{Figure 2}, the ConvMixer block is composed of $l$ ConvMixer layers. A single ConvMixer layer consists of depthwise convolution (kernel size k$\times$k) and pointwise convolution (kernel size 1$\times$1). The number of group channels of the depthwise convolution kernel is equal to the channels of the input feature map. Each convolution is followed by a GELU [39] activation and batch normalization. The ConvMixer module is defined by formula (3) and (4):

\begin{equation}
f_l^\prime=BN\left(\sigma_1\left\{DepthwiseConv\left(f_{l-1}\right)\right\}\right)+f_{l-1}
\label{eq}\end{equation}
\begin{equation}
f_l=BN\left(\sigma_1\left\{PointwiseConv\left(f_l^\prime\right)\right\}\right)
\label{eq}\end{equation}
where $f_l$ represents the output feature map of layer $l$ in the ConvMixer block, $\sigma_1$ represents the GELU activation, and BN represents the batch normalization. It is worth mentioning that the feature maps from all layers in the ConvMixer module maintain the same resolution and size.

{\bf Multi-scale Attention Gate:} Our proposed multi-scale attention gate is depicted in \textcolor{cyan}{Figure 2}. The multi-scale attention gate is integrated with skip-connections, which are used to suppress unimportant features and enhance valuable ones. Specifically, to select encoder features adaptively according to different resolutions, we use three types of convolutions with different receptive fields for feature extraction: pointwise convolution, ordinary convolution (i.e., kernel size 3$\times$3 and stride of 1 and padding of 1) and dilated convolution (kernel size of 3$\times$3, stride of 1, padding of 2 and dilation rate of 2), all three convolutions generate feature maps of the same size. Each convolution has a batch normalization layer, and we concatenate the output feature maps before a ReLU activation. We then select valuable features by another pointwise convolution, as shown in formula (5) and (6):

\begin{equation}
\begin{split}
f_{Concat}=\sigma_{2}(Concat{\{}&BN{\{}PointwiseConv(f){\}}, \\ 
&BN{\{}OrdinaryConv(f){\}},\\
&BN{\{}DilationConv(f){\}}{\}})
\end{split}
\label{eq}\end{equation}
\begin{equation}
f_s= f{\times} \sigma_{3}(PointwiseConv(f_{Concat}))+f
\label{eq}\end{equation}
Where $f$ represents encoding features, $f_{Concat}$ is the concatenated feature, $f_s$ is the output feature from the multi-scale attention gate, and $\sigma_2$ and $\sigma_3$ denotes the ReLU and Sigmoid activation, respectively.

\subsubsection{Multi-Level Global Content Consistency}

To ensure consistency in the predictions of the main decoder and auxiliary decoder for different levels of global context information, we embed shared ConvMixer layers of varying lengths following the encoder and add different noise perturbation settings to the input of the auxiliary encoder, as shown in formula (7) and (8).
\begin{equation}
{\hat{y}}_k=D_{auxk}(f_{l_k}^p), f_{l_k}^p=P_k(f_{l_k})
\label{eq}
\end{equation}
\begin{equation}
\vspace{-0.19cm}
\hat{y}=D_{main}(f_{l_K})
\label{eq}
\end{equation}
where $P_k$ represents the kth different noise perturbation, ${\hat{y}}_k$ and $\hat{y}$  are the predicted outputs of the kth auxiliary decoder and main decoder, respectively. In addition, $f_{l_0}$ represents the encoder output and $f_{l_k}^p$ represents the noise-perturbed version output by the shared ConvMixer layers with length $l_k$. $f_{l_k}^p$ is the input to the kth auxiliary decoder. Moreover, $f_{l_K}$ is the input to the main decoder, which is output by the total-length of ConvMixer module. For each labeled sample ${x_i\in D}_N^l$ and its ground truth $y_i$, the supervised loss $\mathcal{L}_{s}$ is calculated according to formula (9):
\begin{equation}
\mathcal{L}_{s}=\frac{1}{(K+1)\cdot N}\sum_{i=1}^{N}\left[\sum_{k=1}^{K}\mathcal{L}\left({\hat{y}}_{i_k},y_i\right)+\mathcal{L}\left({\hat{y}}_i,y_i\right)\right]
\label{eq}\end{equation}
where K is the number of auxiliary decoders. ${\hat{y}}_{i_k}$ and ${\hat{y}}_i$ represent the ith labeled sample output of the kth auxiliary decoder and the main decoder, respectively. $\mathcal{L}\left(.\right)$ is a standard supervised combination loss, which is defined between the ith prediction ${\hat{y}}_i$ and the ground truth $y_i$ as a combination of binary cross entropy (BCE) and dice loss (Dice) as shown in formula (10):
\begin{equation}
\mathcal{L}\left(.\right)=0.5\cdot BCE\left(\hat{y},y\right)+Dice\left(\hat{y},y\right)
\label{eq}\end{equation}

To leverage valuable knowledge from the unlabeled dataset, we ensure output predictions consistency between the main decoder and auxiliary decoders. Specifically, we calculate and minimize the difference between the main and auxiliary encoders outputs on the unlabeled data $D_{M+A}^u$. The unsupervised loss function $\mathcal{L}_{u}$ is defined as formula (11):
\begin{equation}
\mathcal{L}_{u}=\frac{1}{K\cdot\left(M+A\right)}\sum_{j=1}^{M+A}\sum_{k=1}^{K}\vert\vert\hat{y}_j-\hat{y}_{j_k}\vert\vert
\label{eq}\end{equation}
where ${\hat{y}}_{j_k}$ and ${\hat{y}}_j$ represents the jth unlabeled sample output of the kth auxiliary decoder and the main decoder, respectively.  Then, we optimize the combined loss function $\mathcal{L}_{total}$ to learn from both labeled and unlabeled data as formula (12) :
\begin{equation}
\mathcal{L}_{total}=\mathcal{L}_{s}+\lambda\cdot\mathcal{L}_{u}
\label{eq}\end{equation}
where $\mathcal{L}_{s}$ and $\mathcal{L}_{u}$ are presented in formula (9) and (11), respectively. $\lambda$ is the Gaussian warming up function [33, 34],
$\lambda=w_{max}\cdot e^{(-5{(1-\frac{t}{t_{max}})}^2)}$
, where $w_{max}$ represents regularization weight and $t_{max}$ is the maximal training step.

\section{EXPERIMENT}
\subsection{Datasets}
\subsubsection{BUSI Dataset}
The Breast UltraSound Images (BUSI) [40] open-source dataset consists of 780 breast ultrasound images from 600 female patients, covering 133 normal cases, 487 benign cases and 210 malignant cases with their corresponding ground truths. Following recent studies [26, 43], we utilize all cases and randomly split the BUSI dataset into 70-30 ratios three times (i.e., 526 samples for training and 254 samples for validating) to ensure fair comparison.
\subsubsection{BUS Dataset}
The breast ultrasound (BUS) [41] open-source dataset consists of 562 breast ultrasound images acquired from female patients aged 26 to 78 years, collected by six institutions using five ultrasound devices. BUS includes 306 benign cases and 256 malignant cases, and we use BUS as unlabeled data.
\subsubsection{B Dataset}
The Dataset B (B) [42] consists of 163 breast ultrasound images from multiple female patients collected at the UDIAT Diagnostic Center of Parc Taulí in Sabadell, Spain. The B dataset includes 110 benign cases and 53 malignant cases, and we utilize B as unlabeled data.
\subsubsection{TUS Dataset}
The private Thyroid UltraSound dataset (TUS) was collected using three different ultrasound machines from the Ultrasound Department of the Affiliated Hospital of Qingdao University. It includes 192 cases, with a total of 1942 thyroid ultrasound images and corresponding segmentation results by three experienced radiologists. To ensure fair comparison, we randomly split the TUS dataset into 70-30 ratios three times (i.e., 1359 samples for training and 583 samples for validation).
\subsubsection{TNSCUI2020 Dataset}
The Thyroid Nodule Segmentation and Classification in Ultrasound Images 2020 (TNSCUI2020) [47] public dataset was collected using various ultrasound machines by the Chinese Artificial Intelligence Alliance for Thyroid and Breast Ultrasound (CAAU). The TNSCUI2020 dataset includes 3644 cases of different ages and genders, and we use it as unlabeled data.

\begin{table*}[t]
\caption{Result on Self-Domain. We Report the Mean and Stdev With Three Runs.}
\scriptsize
\renewcommand\arraystretch{1.25}
\label{table}
\begin{center}
\begin{tabular}{cc|c|cccccccc}
\hline
\multicolumn{2}{c|}{\multirow{3}{*}{Method}}                                                                        & \multirow{3}{*}{Venue} & \multicolumn{8}{c}{Metrics(\%)}                                                                                                                                                           \\ \cline{4-11} 
\multicolumn{2}{c|}{}                                                                                               &                        & \multicolumn{4}{c|}{BUSI}                                                                         & \multicolumn{4}{c}{TUS}                                                               \\ \cline{4-11} 
\multicolumn{2}{c|}{}                                                                                               &                        & IoU                 & Recall     & Precision           & \multicolumn{1}{c|}{F1}                  & IoU                 & Recall              & Precision           & F1                  \\ \hline
\multicolumn{1}{c|}{\multirow{2}{*}{\begin{tabular}[c]{@{}c@{}}Fully \\ Supervised\end{tabular}}} & U-Net {[}22{]}   & MICCAI'15              & 69.69±0.95          & 62.84±3.44 & 70.74±3.22          & \multicolumn{1}{c|}{62.96±3.54}          & 82.63±0.23          & 90.27±0.48          & 90.62±0.24          & 89.81±0.05          \\
\multicolumn{1}{c|}{}                                                                             & CMU-Net {[}14{]} & ISBI'23                & 70.81±0.39          & 64.00±2.78 & 72.10±1.76          & \multicolumn{1}{c|}{64.14±2.44}          & 83.04±0.12          & 90.24±0.25          & 90.98±0.22          & 90.08±0.11          \\ \hline
\multicolumn{1}{c|}{\multirow{7}{*}{\begin{tabular}[c]{@{}c@{}}Semi \\ Supervised\end{tabular}}}  & MT {[}33{]}      & NIPS'17                & 65.95±2.33          & 60.58±1.71 & 69.49±3.54          & \multicolumn{1}{c|}{60.31±1.81}          & 80.10±0.61          & 88.26±0.07          & 89.25±0.54          & 87.74±0.35          \\
\multicolumn{1}{c|}{}                                                                             & UA-MT {[}34{]}   & MICCAI'19              & 65.12±1.13          & 61.32±1.33 & 70.27±1.99          & \multicolumn{1}{c|}{60.55±1.34}          & 80.18±0.22          & 88.18±0.14          & 89.44±0.32          & 87.79±0.38          \\
\multicolumn{1}{c|}{}                                                                             & DCT {[}35{]}     & ECCV'18                & 65.90±1.25          & 61.39±2.00 & 69.84±1.38          & \multicolumn{1}{c|}{60.81±2.06}          & 80.25±0.22          & 88.63±0.50          & 89.01±0.78          & 87.91±0.28          \\
\multicolumn{1}{c|}{}                                                                             & CPS {[}37{]}     & CVPR'21                & 65.97±1.46          & 61.06±1.73 & 69.97±0.78          & \multicolumn{1}{c|}{60.59±1.62}          & 80.79±0.57          & 88.94±0.32          & 89.55±0.81          & 88.42±0.49          \\
\multicolumn{1}{c|}{}                                                                             & CTBCT {[}38{]}   & MIDL'22                & 65.89±2.67          & \textbf{62.17±1.71} & 70.41±3.74          & \multicolumn{1}{c|}{61.52±1.95}          & 81.29±0.15          & 89.34±0.32          & 89.83±0.38          & 88.83±0.09          \\
\multicolumn{1}{c|}{}                                                                             & CCT {[}13{]}     & CVPR'20                & 66.16±0.91          & 61.84±1.26 & 69.87±1.97          & \multicolumn{1}{c|}{60.97±1.63}          & 80.49±0.38          & 88.88±0.49          & 89.30±0.89          & 88.14±0.29          \\
\multicolumn{1}{c|}{}                                                                             & MGCC (Ours)      & This paper             & \textbf{68.06±2.45} & 61.49±2.82 & \textbf{73.06±2.63} & \multicolumn{1}{c|}{\textbf{62.53±2.99}} & \textbf{81.45±0.35} & \textbf{89.35±0.96} & \textbf{89.96±0.18} & \textbf{88.98±0.45} \\ \hline
\end{tabular}
\label{tab1}
\end{center}
\vspace{-0.5cm}
\end{table*}

\subsection{Experimental Settings}
\subsubsection{Generative Network Training and Hyperparameter Setting}
Our training methodology is based on the Stable Diffusion model [10], which is divided into two stages. In the first stage, we train the autoencoder using VAE [45]. To generate high-resolution images, the input image is resized to 512$\times$512 and mapped into a 64$\times$64 latent space using the VAE encoder. The latent code is then directly decoded into pixel space, and the Mean Squared Error (MSE) is used as the reconstruction loss to minimize the difference between the input and reconstructed pixel images. In the second stage of training, the pretrained autoencoder (VAE) with frozen weights encodes the pixel image to the latent space, and the latent code is then diffused into Gaussian noise. Specifically, we utilize U-Net [22] for inverse denoising estimation.

In the generating experiment, we use Adam optimizer with a learning rate of 0.000001 for training the autoencoder. The batch size is set to 4, and the training epoch is 1000. For the latent diffusion model, we train it using the AdamW optimizer for 1000 epochs with a learning rate of 0.0001. Additionally, we diffuse Gaussian noise with 1000 steps. To reduce the cost of image generation, we utilize the Denoising Diffusion Implicit Model (DDIM) [46] for generating synthesis samples (t=100 denoising steps).

\subsubsection{Semi-Supervised Segmentation Network Training and Hyperparameter Setting}
To reduce computational costs, we deploy three auxiliary decoder (K=3) and utilize three different noise perturbation functions on the input features of the decoders: F-Noise [13], F-Drop [13] and Dropout [44].

For the semi-supervised segmentation experiment in our work and the comparison method, we use the SGD optimizer with a weight decay of 0.0001 and momentum of 0.9 to train the networks for 300 epochs. The initial learning rate is set to 0.01, and use the poly learning rate strategy to adjust the learning rate. We set the batch size to 8, and each batch consists of 4 labeled and 4 unlabeled images. For the Gaussian warming up function , we follow previous work [33, 38] and set $t_{max}$ to 0.1. Additionally, we resize all images to 256$\times$256 and perform random rotation and flip for data augmentation.

\subsubsection{Evaluation Metrics and Comparison Methods}
We adopt four commonly used metrics, namely Intersection over Union (IoU), Recall, Precision and F1-score, to quantitatively evaluate the performance of different segmentation models. Additionally, we compare our method with seven SSL methods for medical image segmentation, including MeanTeacher (MT) [33], Uncertainty-Aware Mean Teacher (UAMT) [34], Deep Co-Training (DCT) [35], Cross Pseudo Supervision (CPS) [37], Cross Teaching Between CNN and Transformer (CTBCT) [38] and Cross-Consistency training (CCT) [13]. Following previous work [33], we set the decay parameter of exponential moving average (EMA) to 0.9 in MT and UAMT. Moreover, all methods utilize U-Net [22] as the backbone and are implemented on an NVIDIA GeForce RTX4090 GPU using the Pytorch framework for a fair comparison.

\begin{figure}[t]
\begin{minipage}[b]{0.495\linewidth}
  \centering
  \centerline{\includegraphics[width=\columnwidth]{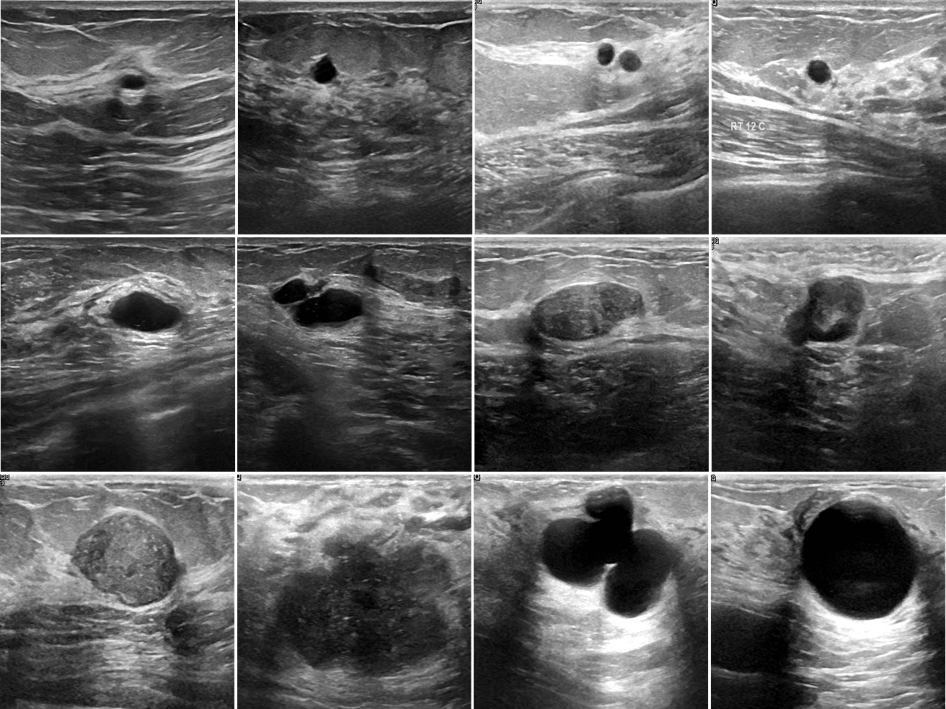}}
  \centerline{(a)}\medskip
\end{minipage}
\begin{minipage}[b]{0.495\linewidth}
  \centering
  \centerline{\includegraphics[width=\columnwidth]{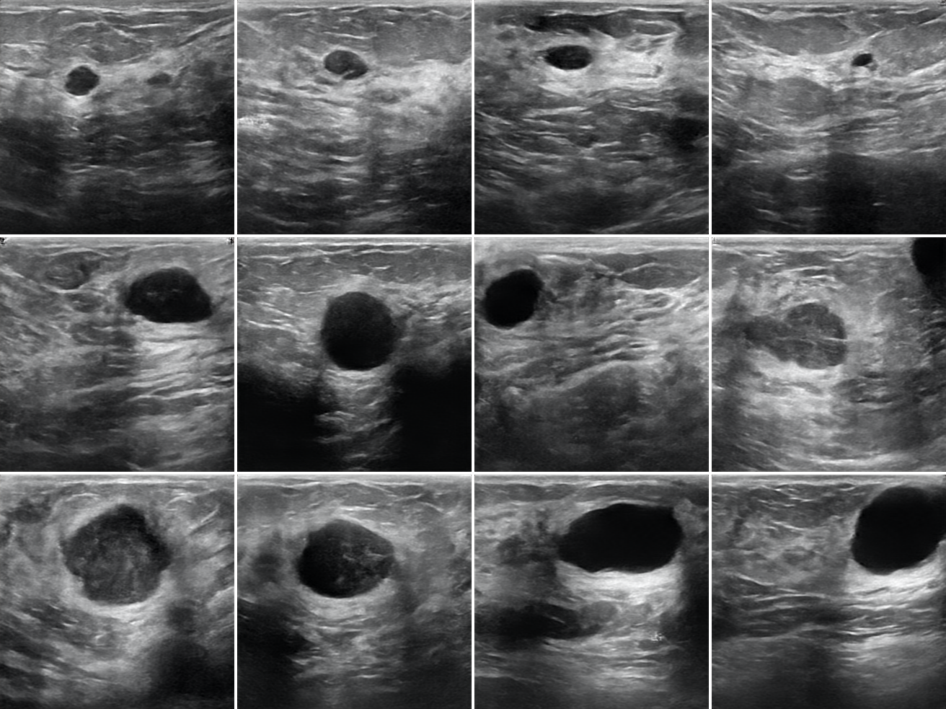}}
  \centerline{(b)}\medskip
\end{minipage}
\begin{minipage}[b]{0.495\linewidth}
  \centering
  \centerline{\includegraphics[width=\columnwidth]{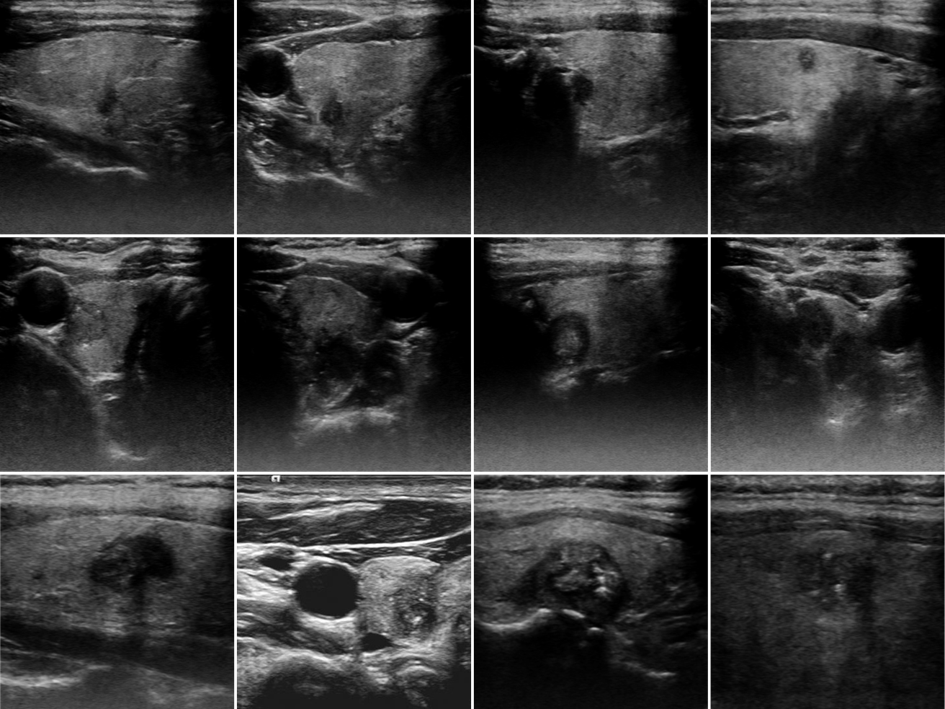}}
  \centerline{(c)}\medskip
\end{minipage}
\begin{minipage}[b]{0.495\linewidth}
  \centering
  \centerline{\includegraphics[width=\columnwidth]{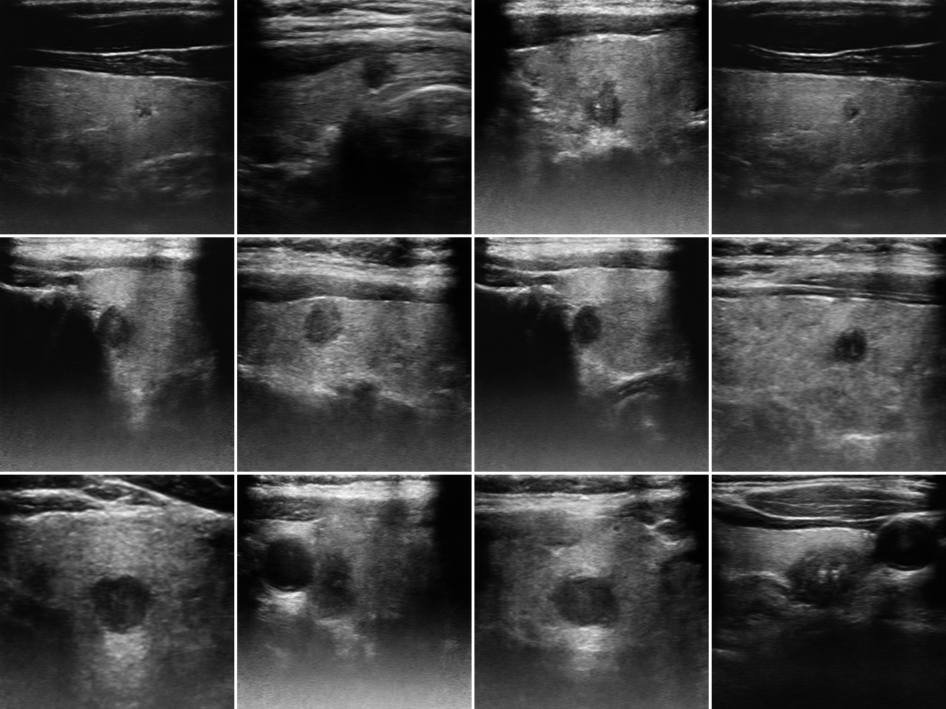}}
  \centerline{(d)}\medskip
\end{minipage}

\caption{Synthetic ultrasound images generated by LDM. The lesion area gradually increases from the first to third rows. (a) Real breast ultrasound images (BUSI).  (b) Synthesis breast ultrasound images. (c) Real thyroid ultrasound images (TUS). (d) Synthesis thyroid ultrasound images.}
\label{fig3}
\vspace{-0.5cm}
\end{figure}

\subsection{Results of Generative Network}
In this experiment, we used LDM to generate images on the BUSI and TUS training sets. We observed a few low-quality synthetic samples in the generated results, we consulted professionals with domain knowledge and selected high-quality synthetic samples (i.e., 725 synthetic samples for BUSI and 3644 samples synthetic of TUS) to create unlabeled synthetic datasets. Some breast and thyroid ultrasound synthetic samples generated by LDM are shown in \textcolor{cyan}{Figure 3}. We found that LDM is capable of generating realistic and diverse breast and thyroid ultrasound synthetic images.

\begin{table*}[t]
\caption{Result on Similar Domains. We Report the Mean and Stdev With Three Runs. $\uparrow$or $\downarrow$ Indicate an Increase or Decrease Relative to the Corresponding Indicators in Table \uppercase\expandafter{\romannumeral1}, and Parentheses Indicate the Diffrernce.}
\scriptsize
\renewcommand\arraystretch{1.25}
\tabcolsep=0.09cm
\label{table}
\begin{center}
\begin{tabular}{cl|c|cccccccc}
\hline
\multicolumn{2}{c|}{\multirow{3}{*}{Method}} & \multirow{3}{*}{Venue} & \multicolumn{8}{c}{Metrics(\%)}                                                                                                                                                           \\ \cline{4-11} 
\multicolumn{2}{c|}{}                        &                        & \multicolumn{4}{c|}{BUSI}                                                                         & \multicolumn{4}{c}{MTS}                                                               \\ \cline{4-11} 
\multicolumn{2}{c|}{}                        &                        & IoU                 & Recall     & Precision           & \multicolumn{1}{c|}{F1}                  & IoU                 & Recall              & Precision           & F1                  \\ \hline
\multicolumn{2}{c|}{MT {[}33{]}}              & NIPS'17                & 65.54±0.96$\downarrow$\tiny(0.41)        &  60.77±0.73$\uparrow$\tiny(0.19)            &          69.25±2.30$\downarrow$\tiny(0.24)           & \multicolumn{1}{c|}{60.13±0.91$\downarrow$\tiny(0.17)}                    &             80.16±0.33$\uparrow$\tiny(0.06)         &       88.68±0.34$\uparrow$\tiny(0.42)              &         89.12±0.60$\downarrow$\tiny(0.12)           &           87.87±0.30$\uparrow$\tiny(0.12)          \\
\multicolumn{2}{c|}{UA-MT {[}34{]}}           & MICCAI'19              & 65.59±1.77$\uparrow$\tiny(0.46)         & 59.86±1.27$\downarrow$\tiny(1.46) & 69.24±2.81$\downarrow$\tiny(1.03)          & \multicolumn{1}{c|}{59.77±1.55$\downarrow$\tiny(0.78)}          & 80.31±0.12$\uparrow$\tiny(0.12)          & 88.44±0.06$\uparrow$\tiny(0.25)          & 89.25±0.30$\downarrow$\tiny(0.19)          & 87.86±0.20$\uparrow$\tiny(0.06)          \\
\multicolumn{2}{c|}{DCT {[}35{]}}             & ECCV'18                & 65.88±1.51$\downarrow$\tiny(0.01)         & 60.87±0.61$\downarrow$\tiny(0.52)  & 69.72±2.42$\downarrow$\tiny(0.12)          & \multicolumn{1}{c|}{60.36±0.96$\downarrow$\tiny(0.45)}          & 80.20±0.19$\downarrow$\tiny(0.05)          & 88.80±0.61$\uparrow$\tiny(0.16)          & 88.95±1.02$\downarrow$\tiny(0.06)          & 87.92±0.19$\uparrow$\tiny(0.01)           \\
\multicolumn{2}{c|}{CPS {[}37{]}}             & CVPR'21                & 66.18±1.05$\uparrow$\tiny(0.21)          & 60.55±1.49$\downarrow$\tiny(0.51)  & 69.31±1.45$\downarrow$\tiny(0.66)          & \multicolumn{1}{c|}{60.30±1.40$\downarrow$\tiny(0.28)}          & 80.35±0.11$\downarrow$\tiny(0.44)          & 88.57±0.14$\downarrow$\tiny(0.37)           & 89.34±0.43$\downarrow$\tiny(0.20)          & 88.07±0.21$\downarrow$\tiny(0.35)          \\
\multicolumn{2}{c|}{CTBCT {[}38{]}}           & MIDL'22                & 65.81±2.05$\downarrow$\tiny(0.07)           & 62.22±0.25$\uparrow$\tiny(0.04) & 70.66±2.28$\uparrow$\tiny(0.25)          & \multicolumn{1}{c|}{61.51±0.65$\downarrow$\tiny(0.01)}          & 80.79±0.20$\downarrow$\tiny(0.50)          & \textbf{89.23±0.35$\downarrow$\tiny(0.10)}           & 89.35±0.12$\downarrow$\tiny(0.48)          & 88.46±0.19$\downarrow$\tiny(0.37)          \\
\multicolumn{2}{c|}{CCT {[}13{]}}             & CVPR'20                & 65.94±1.23$\downarrow$\tiny(0.22)         & 62.02±1.06$\uparrow$\tiny(0.18) & 70.26±1.79$\uparrow$\tiny(0.39)           & \multicolumn{1}{c|}{61.06±1.71$\uparrow$\tiny(0.08)}          & 80.40±0.48$\downarrow$\tiny(0.09)          & 88.62±0.36$\downarrow$\tiny(0.26)          & 89.21±0.36$\downarrow$\tiny(0.09)            & 87.97±0.46$\downarrow$\tiny(0.17)          \\
\multicolumn{2}{c|}{MGCC (Ours)}              & This paper             & \textbf{67.92±1.16$\downarrow$\tiny(0.13)} & \textbf{62.48±1.54$\uparrow$\tiny(0.98)}  & \textbf{72.33±0.87$\downarrow$\tiny(0.72)} & \multicolumn{1}{c|}{\textbf{62.66±1.74$\uparrow$\tiny(0.12)}} & \textbf{81.09±0.42$\downarrow$\tiny(0.35)} & 88.85±0.49$\downarrow$\tiny(0.50) & \textbf{89.80±0.27$\downarrow$\tiny(0.16)} & \textbf{88.56±0.40$\downarrow$\tiny(0.41)} \\ \hline
\end{tabular}
\label{tab1}
\end{center}
\vspace{-0.5cm}
\end{table*}

\begin{table*}[t]
\caption{Result on Synthesis Dataset Domain. We Report the Mean and Stdev With Three Runs. $\uparrow$or $\downarrow$ Indicate an Increase or Decrease Relative To the Corresponding Indicators in Table \uppercase\expandafter{\romannumeral1}, and Parentheses Indicate the Diffrernce.}
\scriptsize
\renewcommand\arraystretch{1.25}
\tabcolsep=0.09cm
\label{table}
\begin{center}
\begin{tabular}{cl|c|cccccccc}
\hline
\multicolumn{2}{c|}{\multirow{3}{*}{Method}} & \multirow{3}{*}{Venue} & \multicolumn{8}{c}{Metrics(\%)}                                                                                                                                                           \\ \cline{4-11} 
\multicolumn{2}{c|}{}                        &                        & \multicolumn{4}{c|}{BUSI}                                                                         & \multicolumn{4}{c}{MTS}                                                               \\ \cline{4-11} 
\multicolumn{2}{c|}{}                        &                        & IoU                 & Recall     & Precision           & \multicolumn{1}{c|}{F1}                  & IoU                 & Recall              & Precision           & F1                  \\ \hline
\multicolumn{2}{c|}{MT {[}33{]}}              & NIPS'17                & 66.22±1.42$\uparrow$\tiny(0.26)          &  61.19±1.70$\uparrow$\tiny(0.61)           &          69.00±2.65$\downarrow$\tiny(0.48)            & \multicolumn{1}{c|}{60.46±1.85$\uparrow$\tiny(0.14)}                    &             80.18±0.06$\uparrow$\tiny(0.08)         &       88.44±0.57$\uparrow$\tiny(0.18)              &         89.04±0.73$\downarrow$\tiny(0.21)            &           87.81±0.20$\uparrow$\tiny(0.06)          \\
\multicolumn{2}{c|}{UA-MT {[}34{]}}           & MICCAI'19              & 66.63±2.24$\uparrow$\tiny(1.51)         & 60.81±1.40$\downarrow$\tiny(0.51) & 69.25±3.03$\downarrow$\tiny(1.02)          & \multicolumn{1}{c|}{60.56±1.50$\uparrow$\tiny(0.01)}          & 80.28±0.22$\uparrow$\tiny(0.09)          & 88.84±0.14$\uparrow$\tiny(0.66)          & 88.89±0.32$\downarrow$\tiny(0.54)          & 87.91±0.38$\uparrow$\tiny(0.12)          \\
\multicolumn{2}{c|}{DCT {[}35{]}}             & ECCV'18                & 67.05±1.35$\uparrow$\tiny(1.15)         & 60.25±1.45$\downarrow$\tiny(1.51)  & 68.32±0.85$\downarrow$\tiny(1.51)          & \multicolumn{1}{c|}{60.22±1.61$\downarrow$\tiny(0.59)}          & 80.31±0.16$\downarrow$\tiny(0.06)          & 88.57±0.52$\uparrow$\tiny(0.06)          & 89.17±0.83$\downarrow$\tiny(0.16)          & 87.91±0.18$\uparrow$\tiny(0.01)           \\
\multicolumn{2}{c|}{CPS {[}37{]}}             & CVPR'21                & 66.25±1.39$\uparrow$\tiny(0.28)          & 59.96±2.72$\downarrow$\tiny(1.10)  & 68.19±0.76$\downarrow$\tiny(1.78)          & \multicolumn{1}{c|}{59.70±2.27$\downarrow$\tiny(0.89)}          & 80.39±0.45$\downarrow$\tiny(0.44)          & 88.31±0.44$\downarrow$\tiny(0.37)           & 89.70±0.72$\downarrow$\tiny(0.20)          & 88.05±0.32$\downarrow$\tiny(0.35)          \\
\multicolumn{2}{c|}{CTBCT {[}38{]}}           & MIDL'22                & 66.56±1.11$\uparrow$\tiny(0.67)           & 62.24±2.14$\uparrow$\tiny(0.06) & 70.13±2.45$\downarrow$\tiny(0.27)          & \multicolumn{1}{c|}{61.60±2.01$\uparrow$\tiny(0.08)}          & 81.00±0.31$\downarrow$\tiny(0.28)          & 88.98±0.32$\downarrow$\tiny(0.36)           & \textbf{89.80±0.39$\downarrow$\tiny(0.03)}          & 88.56±0.27$\downarrow$\tiny(0.27)         \\
\multicolumn{2}{c|}{CCT {[}13{]}}             & CVPR'20                & 65.93±1.58$\downarrow$\tiny(0.23)         & 60.65±2.06$\downarrow$\tiny(1.18) & 69.86±2.05$\downarrow$\tiny(0.01)           & \multicolumn{1}{c|}{60.39±1.75$\downarrow$\tiny(0.58)}          & 80.44±0.43$\downarrow$\tiny(0.05)          & 88.47±0.34$\downarrow$\tiny(0.41)          & 89.35±0.27$\downarrow$\tiny(0.05)            & 87.98±0.34$\downarrow$\tiny(0.15)          \\
\multicolumn{2}{c|}{MGCC (Ours)}              & This paper             & \textbf{68.70±2.58$\uparrow$\tiny(0.64)} & \textbf{62.59±1.72$\uparrow$\tiny(1.09)}  & \textbf{72.08±0.73$\downarrow$\tiny(0.98)} & \multicolumn{1}{c|}{\textbf{62.88±1.70$\uparrow$\tiny(0.34)}} & \textbf{81.30±0.28$\downarrow$\tiny(0.14)} & \textbf{89.32±0.78$\downarrow$\tiny(0.03)} & 89.74±0.66$\downarrow$\tiny(0.22) & \textbf{88.83±0.34$\downarrow$\tiny(0.14)} \\ \hline
\end{tabular}
\label{tab1}
\end{center}
\vspace{-0.5cm}
\end{table*}

\subsection{Results of Semi-Supervised Segmentation on Self-Domain}

In this study, we investigate the impact of using unlabeled samples from the self-domain (i.e., both labeled and unlabeled images are from the same dataset) on segmentation performance. We randomly selected 263 training samples (i.e., 50\%) from the BUSI dataset as the labeled set, and the remaining 263 samples as the unlabeled set. Similarly, we randomly selected 679 training samples (i.e., 50\%) from the TUS dataset as the labeled set, and the remaining 680 samples as the unlabeled set. The segmentation results of BUSI and TUS are presented in \textcolor{cyan}{Table \uppercase\expandafter{\romannumeral1}}. On the BUSI dataset, our proposed method achieves excellent results, with an IoU of 68.06\%, which is 1.9\% to 2.94\% higher than other methods. Furthermore, our method also demonstrates competitive performance compared to fully-supervised methods that use 526 labeled images, with the IoU score being only 1.63\% and 2.75\% lower than U-Net and CMU-Net [14] respectively. In addition, our proposed method (MGCC) also achieved excellent results on the TUS dataset, with IoU and F1-score reaching 81.45\% and 88.98\%, respectively. These results demonstrates that our method is effective in capturing global context information to achieve accurate lesion localization.

\subsection{Results of Semi-Supervised Segmentation on Similar Domains}
In this experiment, we analyzed the impact of using unlabeled samples from a similar domain (i.e., using the B, BUS and TNSCUI2020 datasets as unlabeled datasets for the BUSI and TUS labeled datasets, respectively) on segmentation performance. On the BUSI dataset, we randomly selected 263 of the 526 BUSI training samples (i.e., 50\%) as the labeled set, and added the B dataset (163 images) and the BUS dataset (562 images) to the remaining 263 BUSI images to form an unlabeled set (988 unlabeled images in total). In the TUS experiment, we randomly selected 679 TUS training samples (i.e., 50\%) as the labeled set and combined the remaining 680 TUS images with the TNSCUI2020 dataset (3644 images) as an unlabeled set (4324 unlabeled images in total).

The segmentation results are presented in \textcolor{cyan}{Table \uppercase\expandafter{\romannumeral2}}, and our proposed method achieves the best results on all four metrics for the BUSI and TUS datasets. However, compared to Table I, we observed that the IoU and F1 scores of most methods decreased to varying degrees after adding the similar domain unlabeled datasets (B and BUS datasets or TNSCUI2020 dataset). Our findings suggest that the similar domain has a negative impact on learning of the target domain (BUSI or TUS). Images obtained from different instruments exhibit variations in content and style due to the influence of acquisition operators and instrument parameters. These variations in content and style between different source images widen the domain gap and impede the transferability of knowledge from the B and BUS datasets in the BUSI experiment or the TNSCUI2020 dataset in the TUS experiment.

\begin{table*}[t]
\caption{Result on Synthesis Dataset. We Report the Mean and Stdev With Three Runs.}
\label{table}
\scriptsize
\renewcommand\arraystretch{1.25}
\begin{center}
\begin{tabular}{cc|c|cccccccc}
\hline
\multicolumn{2}{c|}{\multirow{3}{*}{Method}}                                                                        & \multirow{3}{*}{Venue} & \multicolumn{8}{c}{Metrics(\%)}                                                                                                                                                           \\ \cline{4-11} 
\multicolumn{2}{c|}{}                                                                                               &                        & \multicolumn{4}{c|}{BUSI}                                                                         & \multicolumn{4}{c}{TUS}                                                               \\ \cline{4-11} 
\multicolumn{2}{c|}{}                                                                                               &                        & IoU                 & Recall     & Precision           & \multicolumn{1}{c|}{F1}                  & IoU                 & Recall              & Precision           & F1                  \\ \hline
\multicolumn{1}{c|}{\multirow{2}{*}{\begin{tabular}[c]{@{}c@{}}Fully \\ Supervised\end{tabular}}} & U-Net {[}22{]}   & MICCAI'15              & 69.69±0.95          & 62.84±3.44 & 70.74±3.22          & \multicolumn{1}{c|}{62.96±3.54}          & 82.63±0.23          & 90.27±0.48          & 90.62±0.24          & 89.81±0.05          \\
\multicolumn{1}{c|}{}                                                                             & CMU-Net {[}14{]} & ISBI'23                & 70.81±0.39          & 64.00±2.78 & 72.10±1.76          & \multicolumn{1}{c|}{64.14±2.44}          & 83.04±0.12          & 90.24±0.25          & 90.98±0.22          & 90.08±0.11          \\ \hline
\multicolumn{1}{c|}{\multirow{7}{*}{\begin{tabular}[c]{@{}c@{}}Semi \\ Supervised\end{tabular}}}  & MT {[}33{]}      & NIPS'17                & 70.03±1.87          & 62.91±2.26 & 70.35±2.54          & \multicolumn{1}{c|}{62.83±2.70}          & 82.87±0.39          & 90.51±0.70          & 90.72±0.28          & 90.01±0.41          \\
\multicolumn{1}{c|}{}                                                                             & UA-MT {[}34{]}   & MICCAI'19              & 70.18±1.11          & 63.72±1.83 & 71.18±3.27          & \multicolumn{1}{c|}{63.57±2.76}          & 82.77±0.26         & 90.59±0.43          & 90.43±0.35          & 89.90±0.37          \\
\multicolumn{1}{c|}{}                                                                             & DCT {[}35{]}     & ECCV'18                & 69.86±1.23          & 62.09±2.77 & 70.59±2.19          & \multicolumn{1}{c|}{62.49±3.01}          & 82.89±0.34          & 90.57±0.53          & 90.70±0.06          & 90.03±0.26          \\
\multicolumn{1}{c|}{}                                                                             & CPS {[}37{]}     & CVPR'21                & 70.29±0.91          & 62.68±2.28 & 70.45±2.27          & \multicolumn{1}{c|}{62.79±3.13}          & 82.54±0.57          & 90.01±0.32          & 90.87±0.81          & 89.80±0.49          \\
\multicolumn{1}{c|}{}                                                                             & CTBCT {[}38{]}   & MIDL'22                & 70.87±1.57          & 65.01±2.31 & \textbf{71.85±1.87}          & \multicolumn{1}{c|}{64.51±2.17}          & 83.10±0.22          & \textbf{90.75±0.55}          & 90.72±0.30         & 90.20±0.29          \\
\multicolumn{1}{c|}{}                                                                             & CCT {[}13{]}     & CVPR'20                & 70.59±0.34          & 63.19±3.04 & 70.97±2.38          & \multicolumn{1}{c|}{63.39±2.98}          & 82.88±0.10          & 90.70±0.28        & 90.75±0.14         & 90.09±0.08          \\
\multicolumn{1}{c|}{}                                                                             & MGCC (Ours)      & This paper             & \textbf{73.73±1.14} & \textbf{65.08±3.10} & 70.04±1.93 & \multicolumn{1}{c|}{\textbf{64.96±3.32}} & \textbf{83.27±0.32} & 90.67±0.36 & \textbf{90.96±0.49} & \textbf{90.33±0.22} \\ \hline
\end{tabular}
\label{tab1}
\end{center}
\vspace{-0.5cm}
\end{table*}

\subsection{Results of Semi-Supervised Segmentation on Synthesis Dataset}
We investigate the effectiveness of utilizing synthetic samples generated by LDM as an unlabeled dataset for semi-supervised segmentation. In the BUSI and TUS experiment, we randomly selected annotations from 263 BUSI samples (i.e., 50\%) and 679 TUS samples (i.e., 50\%) as the labeled set, respectively. For fair comparison, to form the BUSI unlabeled dataset, we combined 725 synthetic samples generated by LDM with the remaining 263 BUSI images (988 unlabeled images in total). We also added 3644 synthetic samples generated by LDM to the remaining 680 TUS images to form an unlabeled set (4324 unlabeled images in total). \textcolor{cyan}{Table \uppercase\expandafter{\romannumeral3}} presents the comparison results, and our method outperforms other methods for all metrics by a significant margin on both datasets. On the BUSI dataset, compared to the fully supervised model, we achieve an IoU score that is only 0.99\% lower than U-Net. When compared to \textcolor{cyan}{Table \uppercase\expandafter{\romannumeral1}} and \textcolor{cyan}{Table \uppercase\expandafter{\romannumeral2}}, we observe that the IoU scores of most methods have improved to varying degrees. Additionally, for the TUS dataset, compared with \textcolor{cyan}{Table \uppercase\expandafter{\romannumeral1}}, the IoU and F1-score of most methods have improved to varying degrees. However, compared with \textcolor{cyan}{Table \uppercase\expandafter{\romannumeral2}}, the metrics of most methods in the experiment still declined. We believe that this is due to the small ratio of labeled and unlabeled samples (1:7), where labeled samples are flooded by unlabeled samples, leading to overfitting.

Furthermore, we conducted a deeper analysis of the semi-supervised segmentation performance on unlabeled synthetic images. We used all training images of BUSI and TUS as the labeled dataset and the synthesized images as the unlabeled set, respectively. Surprisingly, as shown in \textcolor{cyan}{Table \uppercase\expandafter{\romannumeral4}}, all semi-supervised learning methods achieved higher IoU scores than the fully supervised U-Net. These findings suggest that the segmentation network was capable of effectively utilizing semi-supervised learning to acquire knowledge of diffusion probability in LDM. In addition, compared to fully supervised and semi-supervised learning segmentation methods, our proposed method allows for the maximum possible transfer of diffusion probability knowledge to the model, achieving an IoU of 73.73\% and 83.27\%, and F1-score of 64.96\% and 90.33\% on BUSI and TUS, respectively.

\subsection{Ablation and Comparison Studies}
In this study, we conduct comprehensive ablation and comparison experiments on our proposed method with the BUSI dataset and analysis the contribution of each part of our proposed method. We randomly selected 263 training samples (i.e., 50\%) from the BUSI dataset as the labeled set and the remaining 263 samples as the unlabeled set for semi-supervised learning.
\subsubsection{Ablation Study on Multi-Scale Attention Gates}
In this ablation experiment, we placed the multi-scale attention gate in the skip-connections between different decoders and shared encoders to explore its effect on the segmentation model. The results are shown in \textcolor{cyan}{Table \uppercase\expandafter{\romannumeral5}}, and compared to adding multi-scale attention gates between the partial decoders and the shared encoder or not deploying, deploying multi-scale attention gates between all decoders and the shared encoder achieves higher IoU and F1-score. It shows that the multi-scale attention gates can effectively magnify the influence of helpful encoder features in knowledge transfer.

\subsubsection{Comparison Study on ConvMixer}
The multi-level global context of our proposed method is mainly embodied in the output of different length ConvMixer layers, which are connected with the decoders. With the increasing length of ConvMixer layers, the level of global context information is upgraded. \textcolor{cyan}{Table \uppercase\expandafter{\romannumeral6}} lists the comparative experiments with different decoders connecting with different lengths of ConvMixer layers with various kernel sizes. The results show that connecting auxiliary decoder 2, 3 and the main decoder at the ConvMixer layers with the lengths of 3, 6 and 9, respectively, achieves the best performance.

\begin{table}[t]
\caption{Iou and F1-Value of Our Method With Multi-Scale Attention Gate Ablation Study on Busi Dataset. We Report the Mean and Stdev With Three Runs.}
\label{table}
\scriptsize
\renewcommand\arraystretch{1.25}
\tabcolsep=0.15cm
\begin{center}
\begin{tabular}{cccc|cc}
\hline
\multicolumn{4}{c|}{Multi-scale attention gate location}  & \multicolumn{2}{c}{Metrics(\%)} \\ \hline
\begin{tabular}[c]{@{}c@{}}Aux\\ Decoder 1\end{tabular} & \begin{tabular}[c]{@{}c@{}}Aux\\ Decoder 2\end{tabular} & \begin{tabular}[c]{@{}c@{}}Aux\\ Decoder 3\end{tabular} & \begin{tabular}[c]{@{}c@{}}Main\\ Decoder\end{tabular} & IoU             & F1            \\ \hline
&                                                         &                                                         &                                                        &   67.58±2.22    &    62.42±1.70  \\
&                                                         &                                                         &     $\checkmark$                                                   &    67.72±1.32   &    62.28±2.17  \\
$\checkmark$&      $\checkmark$                                                   &    $\checkmark$                                                     &                                                        &   67.42±1.58    &   62.49±2.84  \\
$\checkmark$&     $\checkmark$                                                    &     $\checkmark$                                                    &   $\checkmark$                                             &   68.06±2.45    &  62.53±2.99     \\\hline
\end{tabular}
\end{center}
\end{table}

\begin{table}[t]
\caption{Iou and F1-Value of Our Method With Convmixer Comparison Study on Busi Dataset.
We Report the Mean and Stdev With Three Runs.}
\label{table}
\scriptsize
\renewcommand\arraystretch{1.25}
\begin{center}
\begin{tabular}{ccc|c|cc}
\hline
\multicolumn{3}{c|}{ConvMixer layer} & \multirow{2}{*}{Kernel Size} & \multicolumn{2}{c}{Metrics(\%)} \\ \cline{1-3} \cline{5-6} 
Layer1     & Layer2     & Layer3     &                              & IoU              & F1           \\ \hline
2          & 4          & 6          & 7                            & 67.38±1.06            & 62.58±2.46      \\
2          & 5          & 9          & 7                            &        66.91±1.71          &      62.75±2.51        \\
4          & 7          & 9          & 7                            &       66.86±0.81           &     62.52±2.49         \\
1          & 4          & 9          & 7                            &     68.01±1.36             &        62.43±1.86      \\
3          & 6          & 9          & 7                            &          68.06±2.45        &      62.53±2.99        \\ \hline
\end{tabular}
\end{center}
\vspace{-0.5cm}
\end{table}

\begin{figure*}[t]
\centering
\begin{minipage}[b]{0.495\linewidth}
  \centering
  \centerline{\includegraphics[width=\columnwidth]{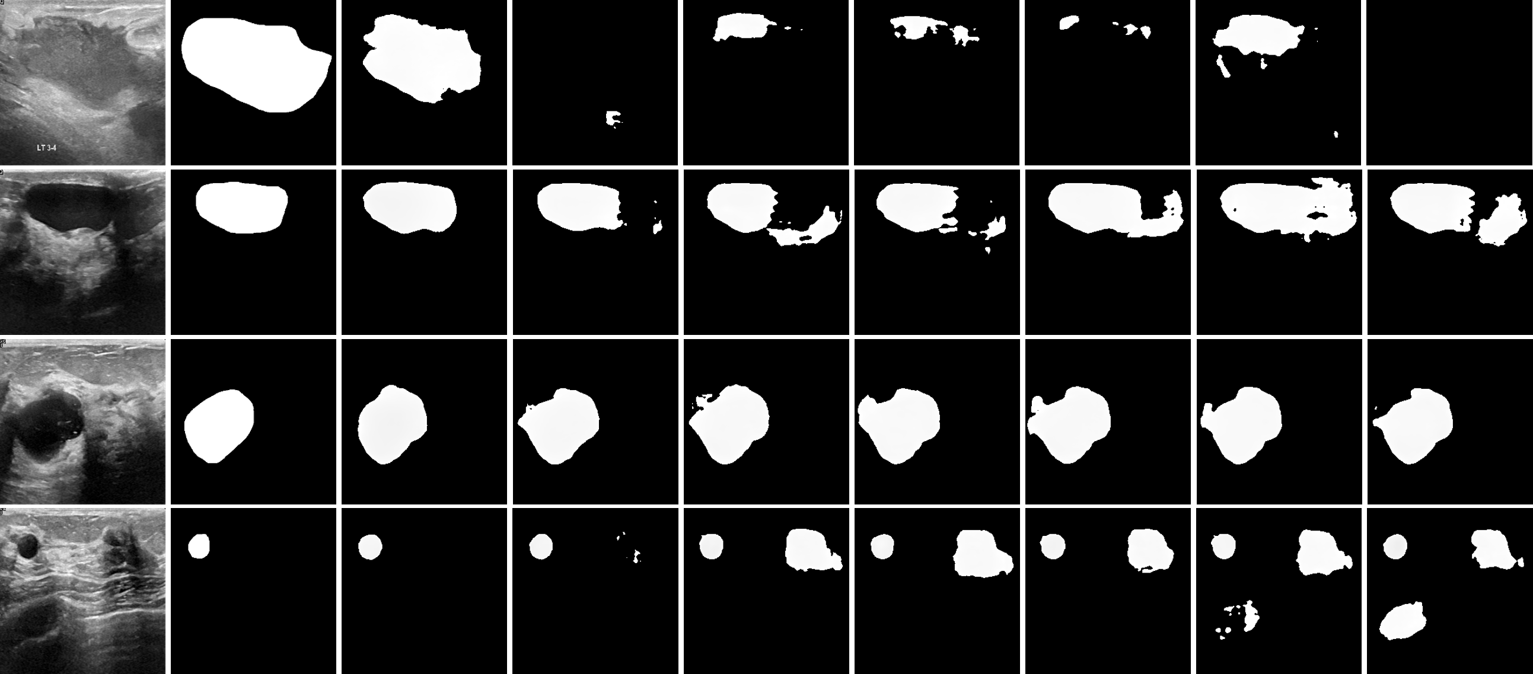}}
  \centerline{\small{\ Input  \ \  \ GT      \  \  \ Ours     \ \ \     MT   \   UA-MT   \  DCT    \ \  CPS  \     CTBCT  \ \ CCT}}
  \centerline{(a)}\medskip
\end{minipage}
\begin{minipage}[b]{0.495\linewidth}
  \centering
  \centerline{\includegraphics[width=\columnwidth]{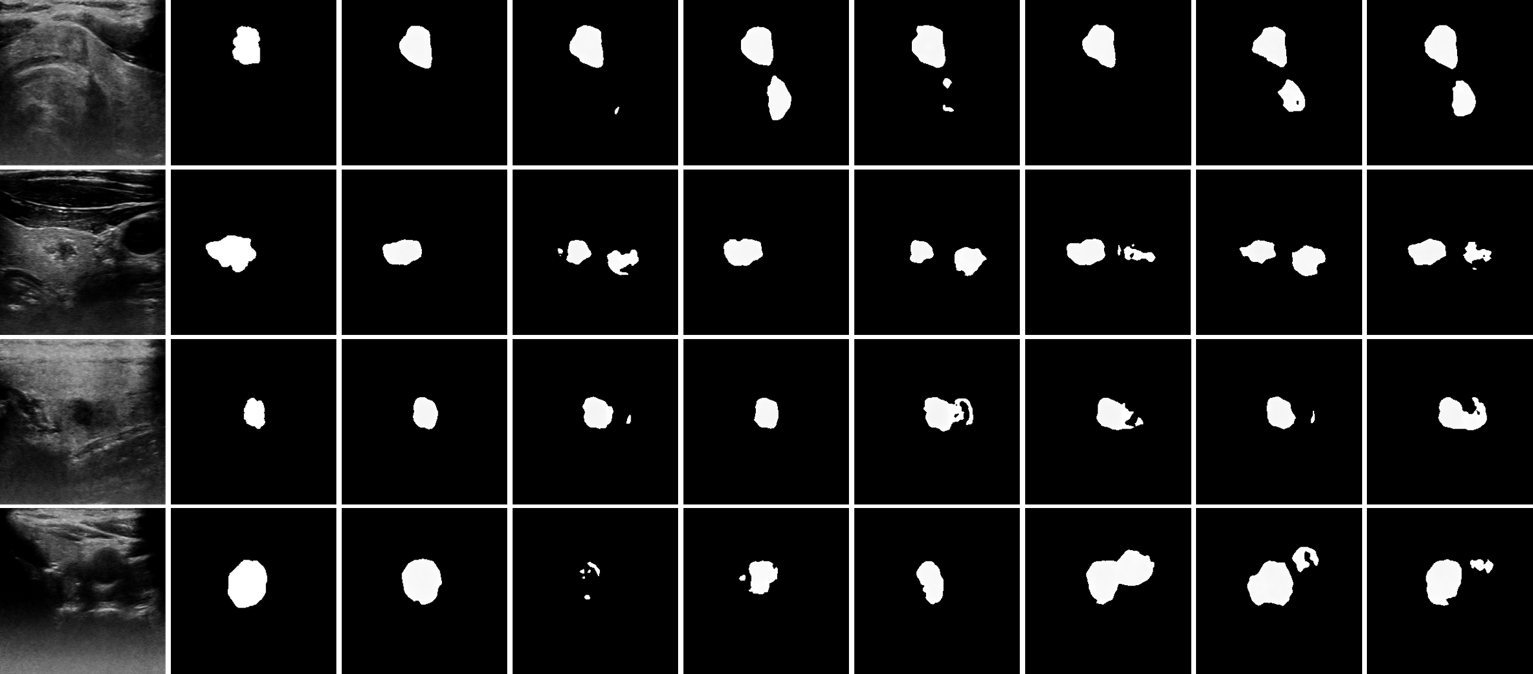}}
  \centerline{\small{\ Input  \ \  \ GT      \  \  \ Ours     \ \ \     MT   \   UA-MT   \  DCT    \ \  CPS  \     CTBCT  \ \ CCT}}
  \centerline{(b)}\medskip
\end{minipage}
\vspace{-0.6cm}
\caption{Visual segmentation result comparison of state-of-the-art methods. (a) BUSI dataset segmentation result. The first row presents challenging segmentation examples with low contrast, the second row presents segmentation examples with unclear echo boundaries, and the third and fourth rows represent segmentation examples for larger and smaller lesions, respectively.  (b) TUS dataset segmentation result.  The first row presents challenging examples which thyroid nodule has blurred edges, the second row presents a lesion with multiple microcalcifications, and the third and fourth rows are examples of larger and smaller lesions.}
\label{fig2}
\end{figure*}

\section{DISCUSSION}
\subsection{Visualization of Results}
We validated the effectiveness of our proposed method by visualizing some segmentation results in \textcolor{cyan}{Figure 4}. On the BUSI dataset, it can be seen that MT, UAMT, DCT, CPS, and CCT have produced poor segmentation. Although CTBCT has improved the results by making use of the global context extracted by Transformer for cross-teaching, it still performs poorly on small targets and generates disconnected regions. In comparison, our method gives more accurate spatial localization and lesion shape. Even for challenging examples with low contrast and unclear echo boundaries, we achieve more complete and convex segmentation results. On the TUS dataset, compared to other methods, our proposed method achieves more complete segmentation results for examples which have blurred edges and multiple microcalcifications. In addition, for lesions at different scales (rows 3 and 4 in \textcolor{cyan}{Figure 4(b)}), other methods have produced poor results. On the contrary, our proposed method achieves more accurate lesion area and shape by learning the global context information.

\subsection{Computational Efficiency and Cost}
The experiment analyzed the average inference time, parameter quantity, and GFLOPs of different methods by using a batch size of 1 testing method. It is worth mentioning that our proposed method does not involve auxiliary encoders during the testing phase, but only includes the main encoder, ConvMixer module, multi-scale attention gates, and main decoder, while other comparison methods are based on U-Net as the benchmark model. The results are shown in \textcolor{cyan}{Table \uppercase\expandafter{\romannumeral7}}. When the network proposed does not add multi-scale attention gates, the inference time and GFLOPs are close to the baseline, and the parameter quantity is 1.28 times larger than the baseline. In addition, when multi-scale attention gates are added, inference time, parameter quantity, and GFLOPs all increase significantly. It can be balanced based on specific actual scenarios between accuracy and inference costs. 

\begin{table}
\caption{Training and Inference Time Of the Tested Methods, For a Batch Size Of 1.}
\label{table}
\label{table}
\scriptsize
\renewcommand\arraystretch{1.25}
\begin{center}
\begin{tabular}{c|ccc}
\hline
Method          & Inference Speed(ms) & Params cost(M) & GFLOPs \\ \hline
U-Net(Baseline) &      20.44               &    34.53            &    65.44    \\
MGCC(w/o msag)  &          22.93           &      44.47          &   67.98     \\
MGCC            &     32.07                &         52.14       &     91.66   \\ \hline
\end{tabular}
\end{center}
\vspace{-0.5cm}
\end{table}

\section{CONCLUSION}
In conclusion, we have proposed a novel multi-level global context cross-consistency (MGCC) framework for medical image segmentation. The framework utilizes the Latent Diffusion Model (LDM) to generate synthetic medical images, reducing the workload of data annotation and addressing privacy concerns associated with collecting medical data. It also includes a fully convolutional semi-supervised segmentation network with multi-level global context cross-consistency, enhancing the network's representational ability for lesions with unfixed positions and significant morphological differences. Through the framework, we successfully leveraged semi-supervised learning to establish a connection between the probability distribution of the target domain and their semantic representations, enabling effective transfer knowledge to the segmentation network.

Experiments on both public and private medical ultrasound image datasets demonstrate the effectiveness of the proposed method. Overall, our proposed method has the potential to be a valuable tool for assisting radiologists in the diagnosis and treatment of breast and thyroid diseases. In the future, our method can be applied to other segmentation tasks to improve segmentation accuracy in small sample scenarios.

\end{document}